\documentclass[letterpaper,10pt,conference]{ieeeconf}

\IEEEoverridecommandlockouts
\usepackage{cite}
\usepackage{graphicx}
\usepackage{times}
\usepackage{amsmath,amssymb,amsfonts}
\usepackage{bm}
\usepackage{booktabs}
\usepackage{multirow}
\usepackage{url}
\usepackage[hidelinks]{hyperref}
\newtheorem{remark}{Remark}
\graphicspath{{figures/}}
\title{\LARGE \bf
AC-SMPC: Differentiable Actor--Critic MPC with State Constraints for Safe and Fast Robot Navigation
}
\author{Chaoran Wu$^{1}$, Jiayu Fan$^{1,\ast}$, Qichen Tan$^{1}$, Yuyi Liu$^{2}$, and Jin Wang$^{1,\ast}$%
\thanks{$\ast$Corresponding authors: jyfan@suda.edu.cn; wjin1985@suda.edu.cn.}%
\thanks{$^{1}$Chaoran Wu, Jiayu Fan, Qichen Tan, and Jin Wang are the School of Future Science and Engineering, Soochow University, Suzhou 215222, China,
        {\tt\small crwu@stu.suda.edu.cn, jyfan@suda.edu.cn, qctan@suda.edu.cn, wjin1985@suda.edu.cn}}%
\thanks{$^{2}$Yuyi Liu is with the Chinese Academy of Sciences Shanghai Institute of Microsystem and Information Technology, Shanghai 200233, China, 
        {\tt\small yuyi.liu@mail.sim.ac.cn}}%
\thanks{This work was supported in part by the Science and Technology Program of Jiangsu Province (BZ2024062), the Natural Science Foundation of the Jiangsu Higher Education Institutions of China (22KJA520007), Suzhou Planning Project of Science and Technology (2023ss03, SYG2025129, SNG2025010), and Key Laboratory of General Artificial Intelligence and Large Models in Provincial Universities, Soochow University.}%
}

\begin{document}

\maketitle
\thispagestyle{empty}
\pagestyle{empty}

\begin{abstract}
Achieving safe navigation in confined dynamic environments remains a challenging problem, as robots must balance navigation efficiency with safety. Existing actor--critic MPC architectures typically employ differentiable MPC layers, but this way cannot readily accommodate explicit state constraints, making it intractable to involve collision avoidance within learning-based planning. To address this issue, we propose an actor--critic model predictive control framework with state constraints. We first use vertices to represent the robot and obstacles, and half-spaces to represent the workspace. Based on these, the nonlinear robot dynamics is locally linearized and the hyperplane-based collision avoidance constraints are convexified into the differentiable MPC layer, and then the actor exploits closed-loop feedback to adjust the MPC objective parameters that shape local motion behavior, while the critic evaluates long-term task performance. We further relax the convexified constraints with bounded and penalized slack variables to form an elastic quadratic program, accommodating cases where the local constraint approximations cannot be satisfied simultaneously. Simulation tests and experiments, involving dynamic collision avoidance for multi-robot in narrow cases, illustrate the advantages of the proposed framework in adaptively adjusting the learned policy and improving navigation efficiency over existing approaches while maintaining the motion safety.
\end{abstract}

\section{Introduction}

\subsection{Motivation}
\label{subsec:motivation}
Fast and safe navigation for robots in confined dynamic environments depends on both interaction timing and geometric clearance. As openings created by moving obstacles change over time, a fixed tracking objective may produce persistent waiting or poorly timed motion. Learning-based MPC combines closed-loop adaptation with model-based prediction~\cite{hewing2020learning,reiter2026survey}. Reinforcement learning can adapt the tracking objective according to observed interactions~\cite{gros2020data,beckenbach2019stage,gao2022adaptation,romero2024acmpc}, while MPC retains the system dynamics, actuator limits, and receding-horizon optimization~\cite{rawlings2017mpc}. Objective adaptation alone, however, does not ensure footprint-level geometric admissibility. A finite collision penalty can discourage proximity but does not impose the corresponding state constraint. Moreover, an inner footprint approximation cannot guarantee that the actual robot body remains collision-free, whereas an outer approximation may introduce conservatism. Explicit geometric constraints can represent collision avoidance and workspace containment directly over the embodied footprint~\cite{zhang2021obca,fan2024trajectory,li2024embodied}, but their nonlinear dependence on robot position and orientation complicates their integration into differentiable actor--critic MPC. We address this problem by locally linearizing the robot dynamics and polygonal geometry and incorporating the resulting separation and containment constraints into a differentiable MPC layer. The actor adapts the tracking objective from closed-loop observations, the critic evaluates long-term performance, and bounded elastic relaxation handles locally incompatible constraint approximations.

\begin{figure*}[t]
    \centering
    \includegraphics[width=\textwidth]
    {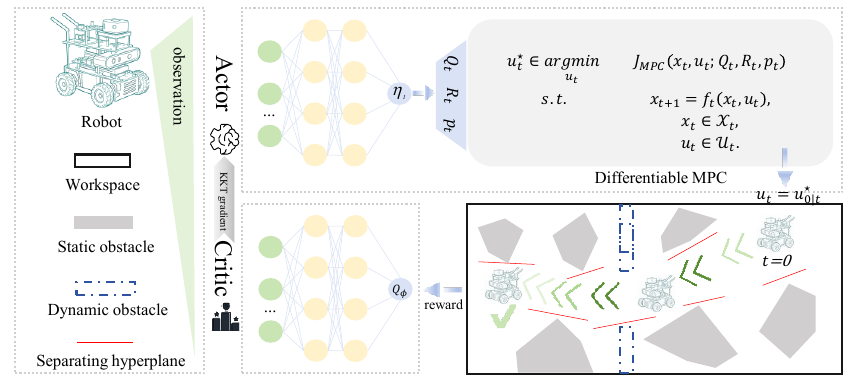}
    \caption{Overview of the AC-SMPC framework. The actor maps the closed-loop observation to bounded adaptation variables, which are decoded into $(Q_t,R_t,\bm p_t)$ for the state-constrained differentiable MPC. The MPC generates the applied control action, while the critic evaluates the resulting interaction and the KKT sensitivity provides the gradient for updating the actor.}
    \label{fig:actmpc_overview}
\end{figure*}

\subsection{Related Work}
\label{subsec:related_work}
Model predictive control is widely used in robotics because it combines prediction, receding-horizon replanning, and explicit input or state constraints in a single online optimization problem~\cite{reiter2026survey}. Reinforcement learning can complement this structure by adapting local motion behavior from task returns rather than relying solely on a hand-designed tracking objective~\cite{gros2020data,beckenbach2019stage,gao2022adaptation,romero2024acmpc}. A central design choice is which part of the predictive controller is learned. Existing approaches learn dynamics, uncertainty descriptions, terminal ingredients, controller approximations, or objective parameters~\cite{berkenkamp2017safe,koller2018safe,rosolia2018learning,kabzan2019racing,hertneck2018approximate}. When the model is available, adapting the tracking objective or controller parameters allows MPC to remain the action-generating layer. Differentiable optimization layers and predictive controllers provide the corresponding gradient path through the optimizer~\cite{donti2017task,amos2017optnet,agrawal2019differentiable,amos2018differentiable,blondel2022implicit,drgona2024constrained,oshin2024differentiable}. In particular, AC-MPC embeds a differentiable MPC layer in an actor--critic policy~\cite{hou2026}, but its action-generating formulation does not include the footprint-dependent separation and containment constraints considered here~\cite{romero2024acmpc}. Objective adaptation can therefore reshape local motion behavior, while footprint-level admissibility requires the corresponding state constraints to be explicitly represented in the action-generating optimization.

State constraints can enter learning-based control at different levels. Constrained and state-wise safe RL incorporate admissibility into policy updates~\cite{achiam2017cpo,chow2018lyapunov,zhao2023statewise}, whereas control barrier functions and predictive safety filters impose model-based conditions on candidate or applied actions~\cite{ames2017cbf,fisac2019safety,wabersich2021predictive,zeng2021dtcbf,jian2023dynamic,brunke2022safe}. These approaches determine how a prescribed admissible set influences learning or control, but its geometric fidelity remains a separate issue. A constraint defined only for a reference point or simplified body may be satisfied even when part of a noncircular robot footprint intersects an obstacle or crosses the workspace boundary.

Optimization-based planners represent collision geometry through signed distances, convex regions, dual formulations, separating planes, and local convex approximations~\cite{thirugnanam2022polytopes,chiu2022collisionfree,guthrie2022distance,zhang2021obca,deits2015iris,marcucci2023convex,fan2024trajectory,Lixiao2025}. Such representations enable the occupied region to be modeled directly rather than inferred from a reference-point trajectory~\cite{li2024embodied}. These works primarily address geometric feasibility under prescribed control objectives. AC-SMPC incorporates this geometric structure into a differentiable actor--critic MPC framework, allowing the tracking objective to be adapted online while the separating and containing constraints remain explicitly represented.

\subsection{Research Objectives and Contributions}
\label{subsec:contribution}
To achieve safe and efficient navigation in confined dynamic environments, we first use geometric representations to model the robot, obstacles, and workspace. A nonconvex obstacle region is constructed as a union of multiple convex polygons, and the workspace is represented as an intersection of half-spaces. The robot footprint and each convex obstacle component are represented as convex polygons. Based on these representations, we introduce two forms of state constraints: separating constraints that prevent collisions between the robot and obstacles, and containing constraints that keep the robot within the workspace. We locally convexify these constraints and embed them into a differentiable MPC layer, thereby explicitly constraining the control action while retaining a local gradient path from the control action to the learned objective parameters. These parameters are adapted online by an actor--critic architecture, in which the actor adjusts the MPC tracking objective from closed-loop feedback and the critic evaluates long-term task performance. The main contributions are as follows.
\begin{enumerate}
\item We propose an AC-SMPC framework which allows involving state constraints in the differentiable MPC layer. Compared to existing penalty-based collision avoidance in the objective function, the proposed framework can improve motion safety for robots.

\item We propose a parameter-adaptive MPC framework by introducing the actor--critic module, which adapts to modify tracking behavior in confined environments. Compared with fixed-cost MPC, AC-SMPC reduces the mean navigation time by 23.1\% (from 12.28~s to 9.44~s), while the mean low-speed ratio decreases from 50.99\% to 18.01\%.

\item We validate the sim-to-real transfer of AC-SMPC for robot navigation in narrow dynamic corridors. The results illustrate the intelligent and fast navigation of AC-SMPC as well as its real-world applicability.
\end{enumerate}

\section{Problem Statement}
\label{sec:problem_statement}

Consider a mobile robot with a polygonal footprint in a confined dynamic environment. The robot must reach a specified goal efficiently while avoiding obstacles and remaining within the bounded workspace. At each closed-loop time $t$, the controller receives the current robot state, a tracking reference, and the predicted obstacle geometry over a finite horizon.

\begin{figure}[t]
    \centering
    \includegraphics[width=\columnwidth]{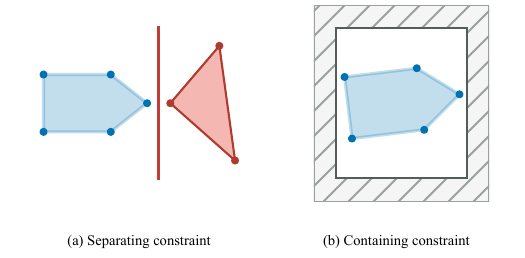}
    \caption{The two studied state constraints. The robot is denoted as a blue polygon. (a) Separating constraint places the robot and obstacles in opposite half-spaces. (b) Containing constraint keeps the robot inside the workspace.}
    \label{fig:geometry_necessity}
\end{figure}

Let $\bm x_t\in\mathcal X\subseteq\mathbb R^{n_x}$ and $\bm u_t\in\mathcal U\subseteq\mathbb R^{n_u}$ denote the robot state and control input at time $t$, respectively. The robot evolves according to the discrete-time dynamics
\begin{equation}
    \bm x_{t+1}=f(\bm x_t,\bm u_t),
    \label{eq:system_dynamics}
\end{equation}
where $f$ is the system dynamics, and $\mathcal{X}$ and $\mathcal{U}$ describe the admissible sets of states and control inputs, respectively.
The symbol $\mathcal B(\bm x)$ denotes the region occupied by the robot footprint at state $\bm x$, $\mathcal O_{k|t}$ denotes the union of the obstacles at prediction stage $k$, and $\mathcal W\subset\mathbb R^2$ denotes the bounded workspace. Obstacle motion is assumed to be known over the finite prediction horizon. A coarse path toward the goal is converted into the horizon tracking reference $\bm r_t$. Let
$\bm X_t=\operatorname{col}(\bm x_{0|t},\ldots,\bm x_{N|t})$ and
$\bm U_t=\operatorname{col}(\bm u_{0|t},\ldots,\bm u_{N-1|t})$
denote the predicted state and input sequences, respectively. Given the tracking-objective parameters $(Q_t,R_t,\bm p_t)$, the resulting finite-horizon navigation problem is formulated as
\begin{equation}
\begin{aligned}
    \min_{\bm X_t,\bm U_t}\quad
    &J_N(\bm X_t,\bm U_t;\bm r_t,Q_t,R_t,\bm p_t)\\
    \mathrm{s.t.}\quad
    &\bm x_{0|t}=\bm x_t,\\
    &\bm x_{k+1|t}=f(\bm x_{k|t},\bm u_{k|t}),
      &&k=0,\ldots,N-1,\\
    &\bm x_{k|t}\in\mathcal X,
      &&k=0,\ldots,N,\\
    &\bm u_{k|t}\in\mathcal U,
      &&k=0,\ldots,N-1,\\
    &\mathcal B(\bm x_{k|t})
      \cap\mathcal O_{k|t}=\emptyset,
      &&k=0,\ldots,N,\\
    &\mathcal B(\bm x_{k|t})\subseteq\mathcal W,
      &&k=0,\ldots,N.
\end{aligned}
\label{eq:footprint_tracking_problem}
\end{equation}
Here, $J_N$ is the tracking objective defined by the state weights $Q_t$, input weights $R_t$, and the reference offset $\bm p_t$. The separating constraints and containing constraints are imposed at every prediction stage $k=0,\ldots,N$. As shown in Fig.~\ref{fig:geometry_necessity}, the separating constraints enforce collision avoidance between the robot and obstacles, while the containing constraints prevent the robot from colliding with the constrained region boundaries. The applied control is the first input of the receding-horizon solution, $\bm u_t=\bm u_{0|t}^{\star}$.

Based on this formulation, we focus on a feedback policy that adapts $(Q_t,R_t,\bm p_t)$ to changing local interactions, as illustrated in Fig.~\ref{fig:actmpc_overview}. These parameters shape the tracking objective, whereas the control action is obtained by solving the constrained problem in~\eqref{eq:footprint_tracking_problem}.

\section{Method}
\label{sec:approach}
\subsection{Geometric Modeling and State Constraints}
\label{subsec:geometric_constraints}
To obtain tractable geometric constraints, we represent the robot footprint and each convex obstacle component as convex polygons and the bounded workspace as a system of linear inequalities. For a convex polygon $\mathcal P$, let
\[
V_{\mathcal P}
=
[\bm v_{1,\mathcal P},\ldots,
 \bm v_{n_v(\mathcal P),\mathcal P}]
\]
denote its vertex matrix. Accordingly, $V_{\mathcal B}(\bm x_{k|t})$ is the vertex matrix of the robot at state $\bm x_{k|t}$, while $V_{\mathcal O_{j,k|t}}$ is the vertex matrix of the $j$th convex obstacle polygon at prediction stage $k$. In this paper, we represent a nonconvex obstacle as a finite union of such convex polygons. The workspace is represented as the intersection of linear half-spaces:
\begin{equation}
    \begin{aligned}
    \mathcal W
    &=\{\bm s\in\mathbb R^2\mid G\bm s\preceq\bm h\}\\
    &=\left\{\bm s\in\mathbb R^2\ \middle|\
    \bm g_l^{\mathsf T}\bm s\leq h_l,\quad
    l=1,\ldots,n_h(\mathcal W)\right\},
    \end{aligned}
    \label{eq:workspace_representation}
\end{equation}
where $\bm g_l^{\mathsf T}$ is the $l$th row of $G$ and $h_l$ is the
corresponding element of $\bm h$. Based on these representations, the state constraints studied in this paper for the embodied footprint are two types of collision avoidance constraints.
\subsubsection{Separating Constraints}

A hyperplane with unit normal $\bm\lambda\in\mathbb R^2$ and offset $\mu\in\mathbb R$ is defined as
\begin{equation}
    \mathcal H(\bm\lambda,\mu)
    =
    \{\bm s\in\mathbb R^2
    \mid \bm\lambda^{\mathsf T}\bm s=\mu\},
    \qquad
    \|\bm\lambda\|_2=1.
    \label{eq:separating_hyperplane}
\end{equation}
By the hyperplane separation theorem~\cite{fan2024trajectory}, the robot and the $j$th convex obstacle polygon $\mathcal O_{j,k|t}$ are separated if there exist $(\bm\lambda_{j,k|t},\mu_{j,k|t})$ satisfying
\begin{equation}
\begin{aligned}
    \bm\lambda_{j,k|t}^{\mathsf T}
    V_{\mathcal B}(\bm x_{k|t})
    &\succeq
    (\mu_{j,k|t}+d_{\mathrm{safety}})\bm 1^{\mathsf T},\\
    \bm\lambda_{j,k|t}^{\mathsf T}
    V_{\mathcal O_{j,k|t}}
    &\preceq
    \mu_{j,k|t}\bm 1^{\mathsf T}.
\end{aligned}
\label{eq:separating_constraints}
\end{equation}
These inequalities place robot and obstacle vertices in opposite half-spaces and prevent interior overlap, where $d_{\mathrm{safety}}$ is added to prevent boundary contact.

\begin{remark}[Separating-hyperplane parameters]
\label{rem:separating_parameters}
The all-ones vectors match the corresponding vertex counts. Let
$\mathcal N_{j,k|t}$ contain both orientations of the unit edge normals of the
nominal robot and obstacle polygons, and let
$\widehat{\bm v}_{i,k|t}$ denote the nominal position of robot vertex $i$.
The separating hyperplane is constructed as
\begin{equation}
\begin{aligned}
\bm\lambda_{j,k|t}
&\in
\operatorname*{arg\,max}_{\bm n\in\mathcal N_{j,k|t}}
\left(
\min_i \bm n^{\mathsf T}\widehat{\bm v}_{i,k|t}
-
\max_\ell
\bm n^{\mathsf T}\bm v_{\ell,\mathcal O_{j,k|t}}
\right),\\
\mu_{j,k|t}
&=
\max_\ell
\bm\lambda_{j,k|t}^{\mathsf T}
\bm v_{\ell,\mathcal O_{j,k|t}}.
\end{aligned}
\label{eq:separating_hyperplane_construction}
\end{equation}
Thus, $\bm\lambda_{j,k|t}$ selects the direction with the largest nominal
projection gap, and $\mu_{j,k|t}$ places the plane on the obstacle support.
The pair is fixed during the QP solve and backward pass and is recomputed after
the nominal trajectory is updated.
\end{remark}

\subsubsection{Containing Constraints}
The containing constraints require every vertex of the robot footprint to satisfy every boundary constraint of the workspace $\mathcal W$~\cite{fan2024trajectory}.
\begin{equation}
    G V_{\mathcal B}(\bm x_{k|t})
    \preceq
    \left(\bm h-d_{\mathrm{safety}}\bm 1\right)
    \bm 1^{\mathsf T},
    \qquad
    k=0,\ldots,N.
    \label{eq:containing_constraints}
\end{equation}
Since $V_{\mathcal B}(\bm x_{k|t})$ varies with the robot position and heading, constraints \eqref{eq:separating_constraints} and \eqref{eq:containing_constraints} are nonlinear and they are locally convexified in the following subsection.

\subsection{Differentiable MPC with State Constraints}
\label{subsec:differentiable_state_constrained_mpc}
To incorporate the state constraints \eqref{eq:separating_constraints} and \eqref{eq:containing_constraints} into the MPC layer, we locally linearize both the robot dynamics and the robot geometry about a nominal trajectory. For clarity, we omit the closed-loop time index from all predicted quantities in the linearization below. The current measured state is denoted by \(\bm x_{\mathrm{cur}}\). A nominal input sequence \(\{\widehat{\bm u}_k\}_{k=0}^{N-1}\) is first constructed, and the corresponding nominal state sequence is generated by a nonlinear rollout:
\begin{equation}
    \widehat{\bm x}_0=\bm x_{\mathrm{cur}},
    \qquad
    \widehat{\bm x}_{k+1}
    =
    f(\widehat{\bm x}_k,\widehat{\bm u}_k).
\end{equation}
The local deviations from the nominal trajectory are defined as
\begin{equation}
    \bm x_k
    =
    \widehat{\bm x}_k+\delta\bm x_k,
    \qquad
    \bm u_k
    =
    \widehat{\bm u}_k+\delta\bm u_k.
    \label{eq:local_increments}
\end{equation}
Linearizing the dynamics along this nominal trajectory gives
\begin{equation}
    \delta\bm x_{k+1}
    =
    A_k\delta\bm x_k
    +
    B_k\delta\bm u_k,
    \qquad
    \delta\bm x_0=\bm 0,
    \label{eq:linearized_dynamics}
\end{equation}
where
\[
    A_k
    =
    \left.
    \frac{\partial f}{\partial \bm x}
    \right|_{(\widehat{\bm x}_k,\widehat{\bm u}_k)},
    \qquad
    B_k
    =
    \left.
    \frac{\partial f}{\partial \bm u}
    \right|_{(\widehat{\bm x}_k,\widehat{\bm u}_k)}.
\]
Collecting the state and input increments over the prediction horizon gives
\[
    \delta\bm X
    =
    [\delta\bm x_1^{\mathsf T},\dots,\delta\bm x_N^{\mathsf T}]^{\mathsf T},
    \qquad
    \delta\bm U
    =
    [\delta\bm u_0^{\mathsf T},\dots,\delta\bm u_{N-1}^{\mathsf T}]^{\mathsf T}.
\]
By recursively substituting \eqref{eq:linearized_dynamics}, the condensed
linearized dynamics are
\begin{equation}
    \delta\bm X=S\delta\bm U,
    \label{eq:condensed_dynamics}
\end{equation}
where $S_0=\bm 0$ and, for $k=1,\ldots,N$, $S_k$ is the block row of $S$
associated with $\delta\bm x_k$.

Since each robot vertex is a function of the robot state, its position is linearized together with the dynamics. For robot vertex \(i\), this yields
\begin{equation}
    \bm v_{i,\mathcal B}(\bm x_k)
    \approx
    \widehat{\bm v}_{i,k}
    +
    J_{i,k}\delta\bm x_k,
    \label{eq:vertex_linearization}
\end{equation}
where
\[
    \widehat{\bm v}_{i,k}
    =
    \bm v_{i,\mathcal B}(\widehat{\bm x}_k),
    \qquad
    J_{i,k}
    =
    \left.
    \frac{\partial \bm v_{i,\mathcal B}}{\partial \bm x}
    \right|_{\widehat{\bm x}_k}.
\]
Substituting the linearized vertex position into \eqref{eq:separating_constraints} and \eqref{eq:containing_constraints}, 
and using the common tightened margin \(m_c=d_{\mathrm{safety}}+m_{\mathrm{lin}}\), where \(d_{\mathrm{safety}}\) specifies the required geometric clearance and \(m_{\mathrm{lin}}\) accounts for the linearization error,
we obtain
\begin{equation}
\begin{aligned}
    -\bm\lambda_{j,k}^{\mathsf T}
    J_{i,k}S_k\delta\bm U
    &\le
    \bm\lambda_{j,k}^{\mathsf T}
    \widehat{\bm v}_{i,k}
    -\mu_{j,k}-m_c,\\
    \bm g_l^{\mathsf T}
    J_{i,k}S_k\delta\bm U
    &\le
    h_l-\bm g_l^{\mathsf T}
    \widehat{\bm v}_{i,k}-m_c,
\end{aligned}
\label{eq:local_geometric_constraints}
\end{equation}
for all relevant vertices \(i\), obstacles \(j\), workspace boundaries \(l\), and prediction stages \(k\).

The state- and input-tracking errors are defined as
\begin{equation}
    \bm e_{k|t}^{x}
    =
    \bm x_{k|t}-\bm r_{k|t}^{x}-\bm p_{k|t}^{x},
    \qquad
    \bm e_{k|t}^{u}
    =
    \bm u_{k|t}-\bm r_{k|t}^{u}-\bm p_{k|t}^{u},
    \label{eq:tracking_errors}
\end{equation}
where \(\bm r_{k|t}^{x}\) and \(\bm r_{k|t}^{u}\) are the state and input components of the horizon tracking reference \(\bm r_t\), and \(\bm p_{k|t}^{x}\) and \(\bm p_{k|t}^{u}\) are the corresponding bias terms decoded from the reference offset \(\bm p_t\). The finite-horizon objective is
\begin{equation}
\begin{aligned}
    J_t
    ={}&
    \frac{1}{2}\sum_{k=0}^{N-1}
    \|\bm e_{k|t}^{x}\|_{Q_{k|t}}^{2}
    +
    \frac{1}{2}\sum_{k=0}^{N-1}
    \|\bm e_{k|t}^{u}\|_{R_{k|t}}^{2}\\
    &+
    \frac{1}{2}
    \|\bm x_{N|t}-\bm r_{N|t}^{x}\|_{P_f}^{2},
\end{aligned}
\label{eq:tracking_objective}
\end{equation}
with \(Q_{k|t}\succeq\bm 0\), \(R_{k|t}\succ\bm 0\), and \(P_f\succeq\bm 0\). Substituting the condensed dynamics into \eqref{eq:tracking_objective} yields a quadratic function of \(\delta\bm U_t\). The local update is obtained from
\begin{equation}
\begin{aligned}
    \min_{\delta\bm U_t,\bm s_t}\quad
    &
    \frac{1}{2}\delta\bm U_t^{\mathsf T}
    H_t\delta\bm U_t
    +
    \bm h_t^{\mathsf T}\delta\bm U_t
    +
    \frac{\rho}{2}\|\bm s_t\|_2^2\\
    \mathrm{s.t.}\quad
    &
    A_t^{s}\delta\bm U_t
    \preceq
    \bm b_t^{s}+\bm s_t,\\
    &
    \bm 0\preceq\bm s_t\preceq\bar{\bm s},\\
    &
    A_t^{u}\delta\bm U_t
    \preceq
    \bm b_t^{u},
\end{aligned}
\label{eq:state_constrained_qp}
\end{equation}
where \(H_t=H_t(Q_t,R_t)\) and \(\bm h_t=\bm h_t(Q_t,R_t,\bm p_t)\),\((A_t^{s},\bm b_t^{s})\) contains the state bounds and the locally approximated state constraints, whereas \((A_t^{u},\bm b_t^{u})\) contains the actuator limits. The bounded slack \(\bm s_t\) relaxes only the state constraints, while the actuator limits remain hard. After each solve, the first optimized input is applied as
\begin{equation}
    \bm u_t
    =
    \widehat{\bm u}_{0|t}
    +
    \delta\bm u_{0|t}^{\star}.
    \label{eq:receding_horizon_action}
\end{equation}

\subsection{Reinforcement Learning for Online Objective Adaptation}
\label{subsec:rl_objective_adaptation}
We employ an actor--critic architecture to adapt the MPC tracking objective online. At each closed-loop step, the actor maps the observation $\bm o_t$ to bounded adaptation variables
\begin{equation}
\bm\eta_t=\pi_{\theta}(\bm o_t)=[\eta_{q,t},\eta_{r,t},\eta_{p,t}]^{\mathsf T},
\qquad
\|\bm\eta_t\|_\infty<0.5.
\label{eq:actor_output}
\end{equation}
where a fixed bound is enforced by a smooth squashing of the raw network output $g_\theta(\bm o_t)$. A differentiable decoder $\mathcal D$ converts $\bm\eta_t$ into the tracking-objective parameters
\begin{equation}
(Q_t,R_t,\bm p_t)=\mathcal D(\bm\eta_t),
\label{eq:objective_decoder}
\end{equation}
preserving the prescribed bounds, $Q_t\succeq0$, and $R_t\succ0$. The state-constrained MPC then applies
\begin{equation}
\bm u_t=\bm u_{0|t}^{\star}(Q_t,R_t,\bm p_t).
\label{eq:mpc_action}
\end{equation}
The critic $Q_\phi(\bm o_t,\widetilde{\bm u}_t)$ evaluates the applied first action, where $\widetilde{\bm u}_t=C_u^{-1}\bm u_{0|t}^{\star}$ is normalized by the diagonal actuator scale $C_u$. With return-to-go $G_t=\sum_{\tau=t}^{T-1}\gamma^{\tau-t}r_\tau$, discount $\gamma\in[0,1]$, and scaled target $\overline G_t=G_t/c_G$ ($c_G>0$), the critic regresses onto the Monte Carlo targets over a finite replay window $\mathcal R$ of recently executed transitions:
\begin{equation}
\mathcal L_Q(\phi)
=\mathbb E_{\mathcal R}\!\left[
\left(Q_\phi(\bm o_t,\widetilde{\bm u}_t^{\mathrm b})-\overline G_t\right)^2
\right],
\label{eq:critic_loss}
\end{equation}
where $\widetilde{\bm u}_t^{\mathrm b}$ is the executed control. 
Let $\bm\xi_t=(Q_t,R_t,\bm p_t)$ and let $\widetilde{\bm u}_t^{\theta}$ denote the normalized first action obtained by re-solving the stored final QP of transition $t$ under $\bm\xi_t=\mathcal D(\pi_\theta(\bm o_t))$, with the constraint data held fixed. The actor loss
\begin{equation}
\mathcal L_\pi(\theta)
=-\mathbb E\!\left[
Q_\phi(\bm o_t,\widetilde{\bm u}_t^{\theta})\right]
+\beta\,\mathbb E\!\left[\|g_\theta(\bm o_t)\|_2^2\right]
\label{eq:actor_loss}
\end{equation}
maximizes the critic value of the re-solved action and regularizes the pre-squashing output; only transitions with a finite re-solved QP and a regular active set contribute to the first term.
Let $\bm\nu_t^\star$ denote the KKT multipliers associated with the active constraints of the final QP. The KKT conditions can then be written as
\begin{equation}
F_{\mathrm{KKT}}(\bm z_t^{\star},\bm\nu_t^{\star};\bm\xi_t)=\bm 0,
\label{eq:kkt_system}
\end{equation}
which yields, by implicit differentiation,
\begin{equation}
\frac{\partial(\bm z_t^{\star},\bm\nu_t^{\star})}
{\partial\bm\xi_t}
=
-
\left(
\frac{\partial F_{\mathrm{KKT}}}
{\partial(\bm z_t^{\star},\bm\nu_t^{\star})}
\right)^{-1}
\frac{\partial F_{\mathrm{KKT}}}{\partial\bm\xi_t}.
\label{eq:kkt_sensitivity}
\end{equation}
Defining $\mathcal L_{\mathrm{KKT},t}=-Q_\phi(\bm o_t,\widetilde{\bm u}_t^{\theta})$, the critic gradient is propagated through the KKT system to the actor:
\begin{equation}
\begin{split}
\frac{\partial\mathcal L_{\mathrm{KKT},t}}{\partial\theta}
&=
\frac{\partial\mathcal L_{\mathrm{KKT},t}}
{\partial\widetilde{\bm u}_t^{\theta}}
\frac{\partial\widetilde{\bm u}_t^{\theta}}
{\partial\bm z_t^{\star}}
\frac{\partial\bm z_t^{\star}}{\partial\bm\xi_t}
\frac{\partial\bm\xi_t}{\partial\theta},
\\
\frac{\partial\bm\xi_t}{\partial\theta}
&=
\frac{\partial\mathcal D}{\partial\bm\eta_t}
\frac{\partial\pi_\theta(\bm o_t)}
{\partial\theta},
\end{split}
\label{eq:actor_gradient}
\end{equation}
where $\partial\bm z_t^{\star}/\partial\bm\xi_t$ is the primal block of \eqref{eq:kkt_sensitivity} and $\partial\widetilde{\bm u}_t^{\theta}/\partial\bm z_t^{\star}$ is the constant Jacobian of the first-control extraction. Because the regularizer lies outside the QP, its gradient is added directly:
\begin{equation}
\nabla_\theta\mathcal L_\pi
=\mathbb E\!\left[\nabla_\theta\mathcal L_{\mathrm{KKT},t}\right]
+\beta\,\nabla_\theta\,
\mathbb E\!\left[\|g_\theta(\bm o_t)\|_2^2\right].
\label{eq:complete_actor_gradient}
\end{equation}
The backward pass differentiates the final stored QP only with respect to $\bm\xi_t$. Under positive definite Hessian, linearly independent active rows, 
and strict complementarity, the solution map is locally differentiable within the active-set region.

\section{Experiments}
\label{sec:experiments}
We conduct simulation and real-world experiments to evaluate AC-SMPC in confined dynamic environments. The simulation studies first examine the effect of online $(Q_t,R_t,\bm p_t)$ adaptation during navigation through dynamic channels. AC-SMPC is then compared with representative real-time approaches and approximate-footprint variants, followed by ablation studies that isolate the roles of the footprint state constraints and elastic relaxation. Finally, physical robot experiments evaluate the deployment of the proposed method during successive dynamic interactions.

In all simulation studies, we use the same discrete-time unicycle model, a prediction horizon of $N=8$, and a control period of $0.20$~s. The limits on linear speed, acceleration, and yaw rate are $2.20$~m/s, $5.00$~m/s$^2$, and $5.00$~rad/s, respectively. $A^*$ provides a coarse path based on the static environment, while moving-obstacle geometry is supplied to AC-SMPC over the prediction horizon.

To train the policy used by AC-SMPC in these experiments, sampled returns are employed as the learning signal. This actor--critic formulation does not require the reward to be differentiated through the MPC layer~\cite{romero2024acmpc}, and we therefore decompose the stage reward into route-progress, elapsed-time, input-variation, and terminal terms as
\begin{equation}
    r_t
    =
    r_t^{\mathrm{prog}}
    +
    r_t^{\mathrm{time}}
    +
    r_t^{\mathrm{smooth}}
    +
    r_t^{\mathrm{term}},
    \label{eq:training_reward}
\end{equation}
where the four terms guide objective adaptation at the task level, while footprint-level geometric admissibility is enforced by the state constraints in the MPC layer.

\begin{figure}[!t]
    \centering
    \includegraphics[width=\columnwidth]
    {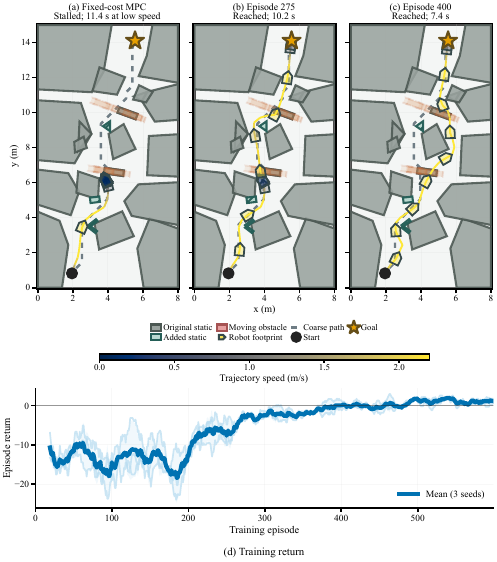}
    \caption{Closed-loop motions and training returns in the confined dynamic environment. Panels~(a)--(c) show fixed-cost MPC and the policies saved at episodes 275 and 400, respectively. The shared legend and color bar define the scene notation and trajectory speed. Panel~(d) shows 20-episode causal moving averages over three seeds; the bold curve and shaded band denote the mean and $\pm$ one sample standard deviation, respectively.}
    \label{fig:training_evolution}
\end{figure}

\begin{figure}[!ht]
    \centering
    \includegraphics[width=\columnwidth]
    {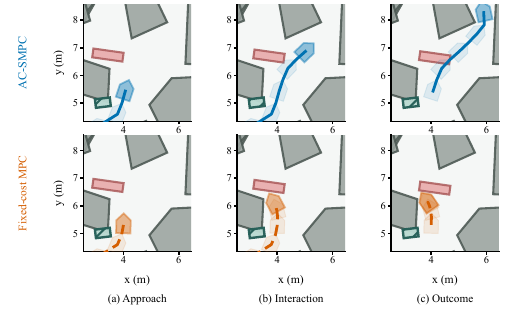}
    \caption{Interaction at the first dynamic passage under identical obstacle motion. The upper and lower rows show AC-SMPC and fixed-cost MPC, respectively, at (a) approaching the obstacle, (b) near the obstacle, (c) the resulting trajectories.}
    \label{fig:dynamic_gate}
\end{figure}
\subsection{Adaptive Navigation in a Confined Dynamic Channel}
\label{subsec:adaptive_navigation}
The scene in Fig.~\ref{fig:training_evolution} contains a pentagonal robot, workspace boundaries, static obstacles, and two rectangular moving obstacles. The moving obstacles follow sinusoidal reciprocating trajectories with a common period of 36 control steps and different amplitudes and phases.

Figure~\ref{fig:training_evolution} shows the evolution of the learned behavior. Fixed-cost MPC stalls before the lower passage and accumulates $11.4$~s of low-speed motion. The policies obtained at episodes 275 and 400 reach the goal in $10.2$ and $7.4$~s, while reducing the low-speed time to $2.4$ and $0.2$~s, respectively. At the first dynamic passage in Fig.~\ref{fig:dynamic_gate}, AC-SMPC enters the opening and continues toward the goal, whereas fixed-cost MPC remains blocked by the moving obstacle.
\begin{figure}[t]
    \centering
    \includegraphics[width=\columnwidth]
    {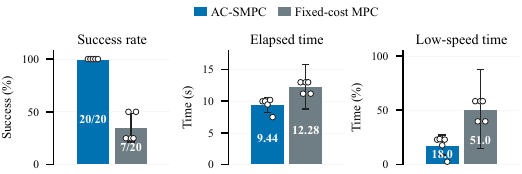}
    \caption{Performance over 20 paired fixed-phase initial-state perturbations. Panels~(a)--(c) report the arrival rate, mean elapsed time, and mean low-speed ratio for $|v|\leq0.32$~m/s, respectively. Dots and error bars denote the mean and one sample standard deviation.}
    \label{fig:held_out_generalization}
\end{figure}
The navigation efficiency gain persists across the 20 paired fixed-phase initial-state perturbations in Fig.~\ref{fig:held_out_generalization}. AC-SMPC increases the mean arrival rate from $35\%$ to $100\%$ and reduces the mean elapsed time from $12.28$~s to $9.44$~s relative to fixed-cost MPC, corresponding to a time reduction of $23.1\%$. The mean low-speed ratio decreases from $50.99\%$ to $18.01\%$. These results show that objective adaptation improves both task completion and navigation efficiency across the tested initial-state perturbations, rather than only in the nominal training trajectory.

\subsection{Comparison with Representative Approaches}
\label{subsec:navigation_performance}
\begin{figure}[!t]
    \centering
    \includegraphics[width=\columnwidth]
    {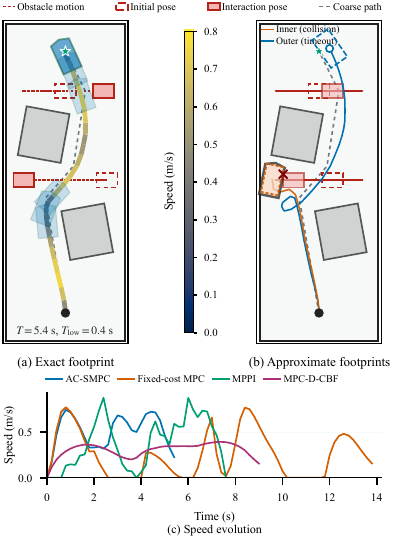}
    \caption{Navigation in the confined dual-dynamic scene. Panel~(a) shows AC-SMPC with the exact footprint. Panel~(b) compares inner and outer rectangular footprint approximations; the orange and blue curves denote their trajectories, and dashed outlines denote the corresponding modeled footprints. The cross marks a collision detected using the exact footprint, whereas the open circle marks the timeout position of the outer approximation. Panel~(c) compares the speed profiles for the configuration in panel~(a).}
    \label{fig:baseline_trajectories}
\end{figure}

We compare AC-SMPC with penalty-only AC-MPC~\cite{romero2024acmpc}, predictive DWA~\cite{fox1997dwa}, MPPI~\cite{williams2017mppi}, RPP~\cite{macenski2023regulated}, and MPC-D-CBF~\cite{jian2023dynamic} in the confined dual-dynamic scene shown in Fig.~\ref{fig:baseline_trajectories}. Four obstacle-phase configurations vary the timing of two successive dynamic interactions. To isolate the effect of footprint representation, we also evaluate AC-SMPC using inner and outer rectangular approximations instead of the exact five-vertex footprint. Penalty-only AC-MPC retains the actor--critic architecture but replaces the state constraints with a collision penalty. All trajectories and geometric violations are evaluated using the exact footprint.

\begin{table}[t]
    \centering
    \caption{Performance over four evaluated phase configurations.}
    \label{tab:advanced_comparison}
    \scriptsize
    \setlength{\tabcolsep}{1.6pt}
    \resizebox{\columnwidth}{!}{%
        \begin{tabular}{@{}lccrrrr@{}}
            \toprule
            Method
            & Arr.
            & Coll.
            & Min clr. (m)
            & Travel (s)
            & Low (s)
            & Path (m) \\
            \midrule
            \textbf{AC-SMPC}
            & \textbf{4/4}
            & \textbf{0/4}
            & \textbf{0.054}
            & \textbf{6.35}
            & \textbf{1.25}
            & 3.044 \\
            AC-SMPC (inner rect.)
            & 1/4
            & 3/4
            & $-0.057$
            & 5.40
            & 0.40
            & 2.762 \\
            AC-SMPC (outer rect.)
            & 2/4
            & 0/4
            & 0.054
            & 7.30
            & 1.90
            & 3.051 \\
            Penalty-only AC-MPC~\cite{romero2024acmpc}
            & 0/4
            & 4/4
            & $-0.001$
            & --
            & --
            & -- \\
            DWA~\cite{fox1997dwa}
            & 3/4
            & 1/4
            & $-0.008$
            & 8.73
            & 4.07
            & \textbf{2.737} \\
            MPPI~\cite{williams2017mppi}
            & \textbf{4/4}
            & \textbf{0/4}
            & 0.021
            & 13.15
            & 8.80
            & 3.085 \\
            RPP~\cite{macenski2023regulated}
            & 0/4
            & 0/4
            & 0.034
            & --
            & --
            & -- \\
            MPC-D-CBF~\cite{jian2023dynamic}
            & 3/4
            & 1/4
            & $-0.017$
            & 8.65
            & 2.55
            & 2.836 \\
            \bottomrule
        \end{tabular}%
    }
    \vspace{1pt}

    \parbox{\columnwidth}{\scriptsize
    Arr./Coll.: arrivals/collisions; Min clr.: minimum signed clearance evaluated using the exact footprint; Low: accumulated time with $|v|\leq0.32$~m/s. Travel, Low, and Path are averaged over successful trials. Bold denotes the best values among methods with $4/4$ arrivals and $0/4$ collisions, except that Path highlights the shortest successful-trial mean among all methods; ``--'' denotes no successful trial.}
\end{table}

Table~\ref{tab:advanced_comparison} shows that AC-SMPC and MPPI are the only approaches that complete all evaluated configurations without collision. Relative to MPPI, AC-SMPC reduces the mean travel time from \(13.15~\mathrm{s}\) to \(6.35~\mathrm{s}\) (\(51.7\%\)) and the mean low-speed time from \(8.80~\mathrm{s}\) to \(1.25~\mathrm{s}\) (\(85.8\%\)), respectively. Its minimum signed clearance also increases from $0.021$~m to $0.054$~m. The shorter travel time is therefore not obtained by accepting a smaller safety margin. DWA and MPC-D-CBF each complete three configurations but collide in the remaining one, penalty-only AC-MPC collides in all four, and RPP reaches none of the goals. Under the joint requirement of completion and collision avoidance, AC-SMPC provides the shortest mean travel time and the largest minimum clearance among the approaches.

The controlled footprint variants explain why the exact representation is needed. The inner rectangle reports a shorter successful-trial time of $5.40$~s, but this value comes from only one arrival; the other three trials collide with the exact footprint and produce a minimum signed clearance of $-0.057$~m. The inner approximation therefore overestimates the available free space rather than improving navigation efficiency. The outer rectangle avoids collision but completes only $2/4$ configurations. In contrast, the exact footprint completes $4/4$ while retaining the same minimum clearance of $0.054$~m, showing that the additional conservatism of the outer approximation does not provide a larger observed safety margin. Figure~\ref{fig:baseline_trajectories} illustrates these mechanisms. Panel~(a) shows a brief slowdown followed by continued progress under the exact footprint, while panel (b) shows the collision incurred by the inner approximation and the stagnation caused by the outer approximation.

\subsection{Ablation Studies}
\label{subsec:ablation_studies}

We evaluate the contributions of the embodied-footprint state constraints and elastic relaxation through the component ablations summarized in Table~\ref{tab:constraint_ablation}.

\begin{table}[!t]
    \centering
    \caption{Ablation results for the proposed components.}
    \label{tab:constraint_ablation}
    \scriptsize
    \setlength{\tabcolsep}{2.2pt}
    \begin{tabular*}{\columnwidth}
        {@{}ll@{\extracolsep{\fill}}ccc@{}}
        \toprule
        Component & Variant & Arrivals & Geom. viol. & QP fail. \\
        \midrule
        \multirow{2}{*}{State constraints}
        & \textbf{Explicit}
        & \textbf{20/20}
        & \textbf{0/20}
        & -- \\
        & Collision penalty
        & 0/20
        & 20/20
        & -- \\
        \cmidrule(lr){1-5}
        \multirow{2}{*}{Elastic relaxation}
        & \textbf{Enabled}
        & \textbf{20/20}
        & \textbf{0/20}
        & \textbf{0/20} \\
        & Removed (hard QP)
        & 6/20
        & 0/20
        & 14/20 \\
        \cmidrule(lr){1-5}
        \multirow{2}{*}{Separation}
        & \textbf{Enabled}
        & \textbf{5/5}
        & \textbf{0/5}
        & -- \\
        & Removed
        & 0/5
        & 5/5
        & -- \\
        \cmidrule(lr){1-5}
        \multirow{2}{*}{Containment}
        & \textbf{Enabled}
        & \textbf{5/5}
        & \textbf{0/5}
        & -- \\
        & Removed
        & 2/5
        & 3/5
        & -- \\
        \bottomrule
    \end{tabular*}
    \vspace{1pt}

    \parbox{\columnwidth}{\scriptsize
    Geom. viol.: collision or boundary crossing;
    QP fail.: QP infeasibility;
    ``--'': not reported.}
\end{table}

Replacing the explicit state constraints with a collision penalty causes all 20 trials to fail, with geometric violations in all 20 trials. Thus, the finite penalty does not provide the same footprint-level admissibility as the explicit constraints in the evaluated trials. The targeted stress tests further distinguish the two studied constraints. Removing separation causes collisions in $5/5$ trials, whereas removing containment causes boundary crossings in $3/5$ trials. These results show that separation and containment address different geometric failure modes and are not interchangeable.

Elastic relaxation affects the solvability of the locally convexified QP. With relaxation enabled, AC-SMPC completes $20/20$ trials without QP failure. Removing it makes the hard QP infeasible in $14/20$ trials and reduces arrivals to $6/20$. The state constraints therefore encode footprint-level geometric admissibility, while elastic relaxation improves the empirical solvability of their local approximations.

\subsection{Real-World Experiments}
\label{subsec:real_world_experiments}

\begin{figure}[t]
    \centering
    \includegraphics[width=\columnwidth]{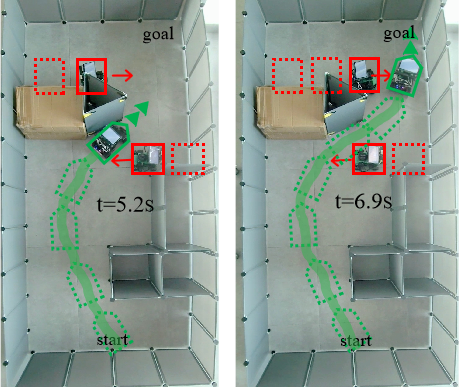}
    \caption{Real-world navigation with two moving obstacles. Green and red denote the AC-SMPC-controlled robot and obstacle robots; solid boxes, dotted outlines, and arrows indicate current footprints, motion histories, and motion directions, respectively.}
    \label{fig:real_world_experiment}
\end{figure}

We implement and evaluate AC-SMPC in a series of real-world experiments using RoboMaster X3 mobile robots. The AC-SMPC-controlled robot is equipped with an NVIDIA Jetson Orin Nano Super, on which actor inference and constrained MPC are executed online. The complete control update, including policy inference and MPC solution, requires $184.6$~ms on average. Two dynamic robots act as moving obstacles, and all robot and obstacle poses are estimated from onboard odometry; obstacle poses are shared via ROS~2. The global path is generated based on a prior static map.

One representative example is shown in Fig.~\ref{fig:real_world_experiment}, highlighting two critical interactions along one representative route. At $t_1=5.2$~s, the observation-conditioned policy coordinates the speed and heading to pass the first moving obstacle and continue directly toward the second passage. At $t_2=6.9$~s, the robot completes the subsequent interaction and reaches the goal. The historical footprints show continuous progression along the route without trajectory accumulation before either opening. These results demonstrate that the learned policy translates local navigation observations into intelligent online interaction strategies, avoiding the prolonged waiting commonly produced by a fixed tracking objective.

\section{Conclusion}
This paper proposes the AC-SMPC for mobile robot navigation in confined dynamic environments. AC-SMPC combines locally convexified separating and containing constraints, elastic QP relaxation, and online objective adaptation to achieve a safe and fast navigation. In the comparisons, the  resulting adaptive behavior translates into faster navigation while retaining positive footprint clearance. Ablation studies verify the distinct roles of the footprint state constraints and elastic relaxation in reducing geometric violations and improving QP solvability, respectively. Real-world experiments further demonstrate the real-time execution through successive dynamic interactions on a physical robot. Future work will address the uncertainty in obstacle-motion prediction and extend the framework to more complex robot dynamics.

\bibliographystyle{IEEEtran}
\bibliography{main}

\end{document}